# USING CONTEXT-DEPENDENT EMBEDDING TO DERIVE FUNDAMENTAL SENTIMENTS AND EXPAND THE SCOPE OF AFFECT CONTROL THEORY

**Moeen Mostafavi & Michael D. Porter**

Systems and Information Engineering

University of Virginia

Charlottesville, VA 22901, USA

`{moeen,mdp2u}@virginia.edu`

**Dawn T. Robinson**

Department of Sociology

University of Georgia

Athens, GA 30602, USA

`{sodawn}@uga.edu`

**Abstract**

Affect control theory (ACT) is a mathematical model of culture-based action that relies on culturally-grounded model specifications to predict behaviors, emotions, and new cultural meanings that arise from local interactions. The theory has been used to make successful predictions about interpersonal social behavior, political actions, social movement strategies, organizational behavior, criminal sentencing, belief transmission, and social emotions. ACT requires dictionaries of social concepts that are indexed in three-dimensions of cultural sentiment. Traditionally, the sentiments are quantified using survey data that is fed into a regression model to explain social behavior. Opportunities to expand the reach of the theory by enlarging the sentiment lexicon are limited due to prohibitive cost. This paper uses a fine-tuned Bidirectional Encoder Representations from Transformers (BERT) model to develop a replacement for these surveys. We compare the performance of this model to that of several alternative embedding techniques. The new model achieves state-of-the-art accuracy in estimating affective meanings, expanding the affective lexicon, and allowing more behaviors to be explained. This model estimates affective meanings similar to running a new survey. This approach could greatly expand the ability of ACT to be applied to new substantive domains at greater speed and at a lower burden for research respondents, researchers, and funders.

# 1 Introduction

Consider talking to your mentor for some advice about how to *behave* with your colleague. Your mentor starts by asking you questions about the *culture* in the workspace and may continue asking about the *identity* of your colleague. These questions might elicit responses about *institutional constraints* such as being the manager, or they may be about *dispositional characteristics* such as being nice or active. Based on this information, your mentor may offer some initial recommendations, but you *adapt* your behavior after observing the reactions from the colleague. This is a descriptive scenario for daily interaction. Affect Control Theory (ACT) is a sociological theory that formalizes the process described in this scenario. This formal theory is both grounded and generative.

The empirically "grounded" part of this theory historically rests on research-intensive data collection methods. While the potential uses of ACT are understanding emotional changes during different social interactions, real life applications are limited due to the vocabulary size of affective dictionaries. Recent advances in computational methodologies offer an opportunity to achieve this grounding in more efficient ways, allowing for more rapid updating to new cultures and cultural changes. Recent work has begun to leverage these methodologies in promising ways. We build on this work and offer a new methodology that improves on the accuracy and expands the capabilities to include new concepts.

In this section, we describe the ACT framework and its limitations due to data collection challenges. Then we discuss computational solutions introduced to overcome its limitation. Finally, we present our computational methodology, a pre-trained deep neural network to expand affective lexicons.

## 1.1 Affect Control Theory

ACT is a mathematical model of culture-based action that uses a set of event-processing equations to describe how affective meanings shift during social interactions. The theory uses culturally-grounded model specifications to predict behaviors, emotions, and new cultural meanings that arise from local interactions. There are three core components of the theory – a scheme for representing *cultural sentiments*, a system of *event processing* equations that characterize cultural rules, and a *control system* logic that animates the meaning-preserving assumption of the theory and makes the theory generative.

**Cultural Sentiments.** The theory represents cultural sentiments in a three-dimensional affective space (Heise, 1977). ACT uses *Evaluation* [good vs. bad], *Potency* [powerful vs. powerless], and *Activity* [active vs. pas-sive] (EPA) space introduced by Osgood et al. (1957) to index social concepts (identities, behaviors, emotion, settings, etc.) in affective space and place them all in a common metric. The EPA dimensions describe substantial variation in the affective meaning of lexicons in more than 20 national cultures studied (Osgood et al., 1975; 1957). Fontaine et al. (2007) found that EPA scores represent the first three principal components after reducing dimensionality on 144 features representing the main components of emotions.

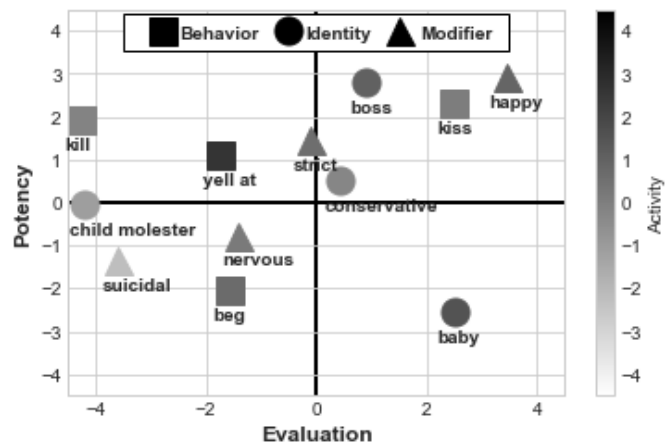

Figure 1: Sentiments of sample words in EPA space. Circles, squares, and triangles show identities, behaviors, and modifiers. The color represents the activity dimension.

ACT considers social interactions or events that include an *actor* that *behaves* toward an *object*. Extracting the Actor-Behavior-Object (ABO) components of an event is the first step in modeling interactions (Heise, 2010). Actor/Object has an identity such as "baby" or "boss" that is represented in the affective space. In some cases, the characteristics of an actor/object is part of the identity. For example, the identity of a person is "nervous boss". In these cases, modifiers (e.g., nervous) are amalgamated to their identities.

Figure 1 is a visualization of U.S. cultural sentiments toward twelve concepts in EPA space. In this plot, we can observe that "suicidal" and "nervous" both have bad, powerless, and passive meanings but "suicidal" is more negative in all three dimensions. On the other hand, "happy" is a pleasant, powerful, and mildly active concept. Note that the range of EPA ratings is from -4.3 to 4.3.

**Impression change equation** People label elements of their interactions (events), and those labels evoke meaning as indexed in the EPA sentiment dictionaries. Consider an example of observing "a bossy employer arguing with an employee". This observation leads people to evaluate both the *actor* and the *object* of the interaction as less pleasant than initially thought. After observing this event, they may also feel the employer is more powerful, and the employee is more powerless than their baseline sentiments. Being more pleasant/powerful is the *impression* of observing this event and it translates to a higher value in *evaluation/potency*. These transient meanings are labeled "impressions" in ACT. Impressions are contextualized affective meanings evoked by symbolic labels in specific social events.

ACT impression change equations predict how sentiments combine to form impressions. These equations are grounded with data from respondents within a language culture. The core impression change equations describe event dynamics from basic interactions of the form "the actor behaves toward the object-person" (called ABO events). ACT also has impression change equations to predict emotional responses to events (Averett and Heise, 1987)), interactions between emotions and identities on situational meanings (Heise and Thomas, 1989; Smith, 2002), non-verbal behaviors (Rashotte, 2002)), self-directed behaviors (Britt and Heise, 1992; Smith and Francis, 2005) , and social settings (Smith, 2002; Smith-Lovin, 1979).

**Control System.** If the *actor* behaves as expected, then the *impression* of their identity will not change far from baseline, but if the *actor* does something unexpected, then a large change from the baseline is expected. *Deflection* is the Euclidean distance between the baseline sentiments of an ABO characters and their impressions following an event. If the impression of an ABO event is close to the initial sentiment, *deflection* is small, but grows bigger as the impression of the event drifts from the initial sentiment. ACT theory suggests that minimizing *deflection* is the driving force in human activity. Highly deflecting events create social and physiological distress (Goldstein, 1989). For example, if a grandmother fights with her grandchild, the grandmother and the grandchild feel distressed. They prefer to do something to bring the impression of their identities back to where they view themselves in the society.For example, we may expect one side to take an action, like apologize. This highly deflected event is very different from two soldiers fighting in a battle. The soldiers are supposed to fight with enemies in battle, so they may not feel social pressure to change their behavior.

The ACT impression change equations are mathematically manipulated to implement the *affect control principle*: the assumption that individuals behave to maintain or restore cultural affective meanings associated with activated labels. The inputs to the impression change equations are EPA sentiment profiles – the culturally shared, fundamental meanings that people associate with social labels. The outputs of those equations, impressions, are transient meanings that arise as social interactions unfold. Discrepancies between sentiments and impressions signal how closely interactions are confirming to cultural prescriptions. The control system part of affect control theory models the assumption that social actors try to maintain their cultural definitions of social situations (the affect control principle). When impressions vary from sentiments (as the temperature might vary in a room), people behave socially to bring the impressions back in line with cultural sentiments. Affect control theorists define deflection as the discrepancy between fundamental cultural sentiments and transient situated impressions. Deflection is the error signal, operationalized as the squared Euclidian distance in EPA space, between cultural sentiments and event impressions. After an event that has disturbed meanings, solving for the behavior profile produces the creative response that an actor is expected to generate to repair the situation. Alternatively, these same equations can predict a new normative understanding that makes sense of the observed events.

Heise (2013) developed a software called INTERACT that can be used to simulate interactions. It uses the impression change equations to solve for behaviors that minimize the deflection or predict attributes and emotions during the interaction. Using INTERACT researchers have access to the theory and use it principally and rigorously without needing to master the technical details. This tool is available and interactive using natural language input. So they can use the theory in a reproducible way without needing to derive the math all over. Consider the following set of events/interactions that we simulate using INTERACT,

1. Employee greets bossy employer.
2. Bossy employer asks employee.
3. Employee replies to bossy employer.
4. Bossy employer argues with employee.
5. Employee listens to / disobeys bossy employer.

The visualization in Figure 2 shows how the impression of actor/object's identity changed based on the sequence of interactions. Let's focus on the evaluation dimension for the employer. The employer has a negative baseline evaluation, but it increases after observing the first two interactions. The first two interactions include positively rated behaviors. After the second interaction, the impression of the employer's identity is positively evaluated and so the next positively evaluated action, replies to, does not move it substantially. A positive behavior is expected from a positive identity. However, in the fourth interaction, the employer is evaluated to have an unpleasant identity after doing a negative behavior, argue with. For the fifth event, we have shown how the impression of different actions by the employee has significantly moved the states for both the actor and the object. The sequential interactions discussed here are similar to our mentorship example discussed earlier. It shows how understanding the interaction dynamic can help predict the consequences of behaviors.

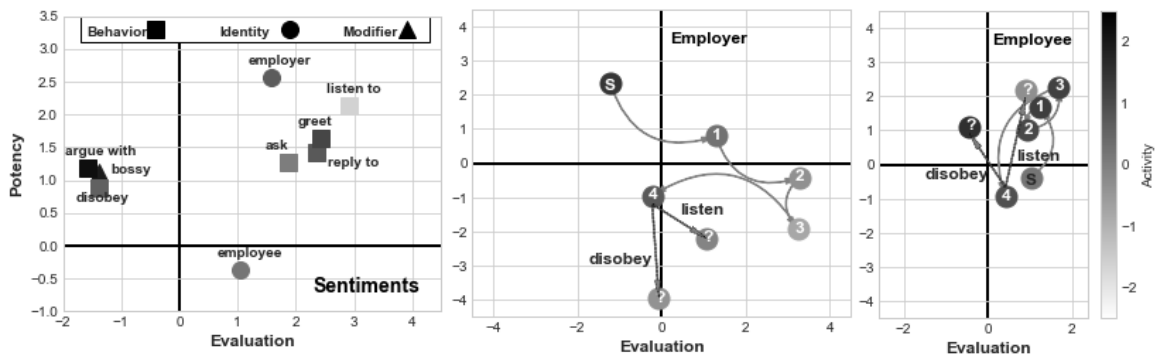

Figure 2: Simulating sequential events in an interaction between an employee and employer. The initial sentiments for both characters are shown by s and the sentiments after each of four events are shown by numbers. After the fourth event, based on what the employee does, the final sentiment for the two characters would be one of the places shown by the question mark.

ACT has rules to describe how the impression of an event changes affective meaning of ABO characters. ACT uses either mathematical equations or descriptive forms to discuss these rules. The following two descriptive forms show how the identity of the *actor* is impressed by some events,

- *Actors* seem nice when they behave in a positive way toward others. This describes *morality* fact in ACT literature. Observing the *evaluation* dimension for the *actors* after he greets the *object*, we can find this *behavior* resulted in an impression of being nicer (getting larger evaluation) comparing to the state in the last step.

- Active *behaviors* make the *actors* seem more active. Observing the boss's activity, he is considered more active after he argues with [active behavior] the employee.

As we have seen in the descriptive forms, events can change the impression of ABO characters. They move them toward or away from their current or initial sentiments.

**Empirical testing and application.** ACT was introduced in the 1970s (Heise, 1977). Validations and applications of affect control theory appear in more than a hundred published articles, book chapters, and books (Robinson and Smith-Lovin, 2018; MacKinnon and Robinson, 2014). Observational studies have revealed dynamics predicted by affect control theory equations (e.g., Francis (1997a;b); Hunt (2008) ). Experimental studies have validated the affect control theory predictions about emotional experiences (e.g., Robinson and Smith-Lovin (1992); Robinson et al. (2012), about identity attributions (e.g., Heise and Calhan (1995); Robinson and Smith-Lovin (1999); Robinson et al. (1994); Tsoudis and Smith-Lovin (1998)) , as well as its predictions about social behavior and performance (Robinson and Smith-Lovin, 1992; Schröder and Scholl, 2009; Youngreen et al., 2009). Surveys studies have validated the theory's predictions about subcultural variation (e.g. Smith-Lovin and Douglass (1992) ). ACT has been used in interdisciplinary applications such as Human-Computer Interactions (Robillard and Hoey, 2018), finding how language cultures affect social response (Kriegel et al., 2017), and modeling identities and behaviors within groups (Rogers and Smith-Lovin, 2019). More recently, Mostafavi (2021) introduced ACT to estimate and track emotional states during online messaging. For example, chatbots can use the ACT framework to understand the emotional state of the customer in real time and adapt their behavior accordingly.

**Empirical grounding.** To formulate the process in mathematically, we briefly review the quantification process using surveys. The first step is quantifying the *sentiments* that are introduced as *identity, behavior, and modifier*. For this purpose, at least 25 participants rate words of interest in EPA space (Heise, 2010). In this survey, the participants rate how they feel about *identity/behavior/modifier* such as "employer".

The sentiment ratings for each concept are aggregated and indexed in a sentiment "dictionary" for each language culture. These supply the baseline meanings for the ACT model. The next step is identifying the event processing dynamics that lead to social impressions (Robinson and Smith-Lovin, 1999). For this purpose, research participants rate ABO characters again after observing a set of events. For example, participants rate affective meaning of "employee", "greet", and "employer" after observing "the employee greets the employer". As we discussed earlier, the ratings of ABO (impressions) could be different from the initial baselines (sentiments). ACT uses regression models, known as impression change equations, to estimate these changes (Heise, 2013). Let $X = [A_e, A_p, A_a, B_e, B_p, B_a, O_e, O_p, O_a]^T$ represent the EPA values/sentiments of an ABO triple, where $\{A, B, O\}$ represent the ABO characters and $\{e, p, a\}$ the EPA components. Consider further the two-way interactions $X^2 = [A_eB_e, A_eO_e, A_eB_a, \ldots A_aO_a]^T$ and three-way interactions $X^3 = [A_eB_eO_e, A_eB_eO_p, A_eB_eO_a, \ldots A_aB_aO_a]^T$. The basic structure of an impression change equation is the linear model

$$X' = \alpha X + \beta X^2 + \gamma X^3 \tag{1}$$

where $\alpha, \beta$, and $\gamma$ are coefficient vectors and $X'$ represent the resulting impression after the event. Modifiers can incorporated prior to impression change by changing the baseline values/sentiments (e.g. bossy employer). Averett and Heise (1987) defined *amalgamation equations* similar to (2) to find the sentiments for an identity with a modifier.

$$A = \rho + \theta M + \psi I, \tag{2}$$

$$\rho = \begin{bmatrix} -0.17 \\ -0.18 \\ 0 \end{bmatrix}, \quad \theta = \begin{bmatrix} 0.62 & -0.14 & -0.18 \\ -0.11 & 0.63 & 0 \\ 0 & 0 & 0.61 \end{bmatrix}, \quad \psi = \begin{bmatrix} 0.50 & 0 & 0 \\ 0 & 0.56 & 0.07 \\ 0 & -0.05 & 0.60 \end{bmatrix}.$$

where, $A$, $M$, $I$, represent amalgamated affective meaning of actor's identity, modifier, baseline identity and $\rho, \theta, \psi$ are the vectors of intercepts and coefficient matrices. Equation (2) is a weighted average of the affective meaning for the modifier and the identity.

## 1.2 Challenges

Current state-of-the-art practices for generating affect control theory models for new cultures require (a) assessing the meanings of 1000-2,000 commonly used labels for social event descriptors (*behaviors, identities, settings, emotions*) in three dimensions of meaning (*evaluation, potency, activity*) and (b) conducting a 512-condition (partial repeated measures) survey experiment to generate data for the event-processing models (Heise, 2010; Kriegel et al., 2017; Rogers, 2018). These techniques are relatively efficient – requiring input from only a few thousand people to produce robust, generative models capable of predicting millions of events. The impression change equations are relatively stable across time, but the sentiment dictionaries vary more in response to social change and by subculture (Heise, 2010). So, updating sentiment dictionaries is a relatively efficient way to tune the theory to a new social context or time. Nonetheless, collecting a new sentiment dictionary in advance of every investigation into a new substantive domain reduces the utility of the theory. To compensate for measurement error between respondents, most EPA surveys are designed so that each word is scored by at least 25 different participants (culture experts). Thus, finding the affective meaning for 5000 words requires over 125,000 ratings and 400 hours of respondent time (Heise 2010). Due to the high cost and time required, most EPA data collections have been limited to relatively small (1-2k words) dictionaries which has limited the applicability of ACT tools.

## 1.3 Toward a computational solution

The ACT usage can be hammered by the lack of access to the concept that needed for the application of interest. Developing a tool that expands concepts used in a dictionary can break limits for researchers.

As an alternative to conducting surveys for a large set of new concepts is expanding the current dictionaries to include those concepts. Researchers have tried supervised (Mostafavi and Porter, 2021; Li et al., 2017) and semi-supervised methods (Alhothali and Hoey, 2017) on shallow word-embeddings to build affective lexicons. Shallow word-embeddings use a neural network with a few layers to find high dimensional vectors representing words or tokens. For example, Mikolov et al. (2013) introduced "Word2Vec" which has 300-dimensional vectors corresponding to 3 million words and phrases. To find these vectors that are called embedding, they trained the neural network on very large corpus such as Google News dataset, which includes about 100 billion words. We review these early practices to expand dictionaries and discuss their limitation. In a recent paper, van Loon and Freese (2022) point to a potential direction for addressing some limitations of shallow-embedding. We review their work that uses Bidirectional Encoder Representations from Transformers (BERT) as a contextual word-embedding alongside shallow word-embedding to estimate affective meaning of words. We share the intuitions expressed by van Loon and Freese (2022), that BERT model might "bridge the gap between the 'words' of embedding models and the 'concepts' pursued by ACT".

Alhothali and Hoey (2017) used graph-based sentiment lexicon induction methods to find affective sentiments associated with words. Knowing the affective meaning of some seed words, the algorithm estimates the affective meaning of new words by propagating the meaning from neighbor words. Their label propagation algorithm finds a similarity measure between a large set of words. In this algorithm, the weights used to find the meaning of a new word are estimated from the similarity of the new word with words in affective dictionary. For example, to estimate affective meaning of *dean* we may look at words such as *boss*, *professor*, and *faculty* in affective dictionaries, and estimate affective meaning of *boss* using a weighted mean of values for known words. Alhothali and Hoey (2017) used similarity graphs to expand affective meanings to neighbor words in four different embedding spaces: (1) semantic lexicon-based Label propagation, (2) distributional based approach which uses co-occurrence metrics in a corpus, (3) neural word embeddings method, and (4) combination of semantic and distributional methods. They found that using both semantics and distributional-based approaches gave the best semi-supervised result. On the other hand, Li et al. (2017) argued that word-embedding can represent the words' general meaning, including denotative meaning, connotative meaning, social meaning, affective meaning, reflected meaning, collocative meaning, and thematic meaning. So, word-embedding graph propagation that uses general meaning similarities may reduce accuracy for finding affective meaning.

Mostafavi and Porter (2021); Li et al. (2017) used supervised methods on shallow word-embeddings to find affective meanings of the words. To put it simply, they use supervised methods to find a mapping from higher dimensional embedding to affective space. For example, Li et al. (2017) use a regression model to estimate the three-dimensional affective meaning from a 300-dimensional "Word2Vec" embedding. All three of these computational approaches described here yielded good results for deriving *evaluation* sentiments for new concepts, but their performance on the *activity* and *potency* dimensions was not very strong.

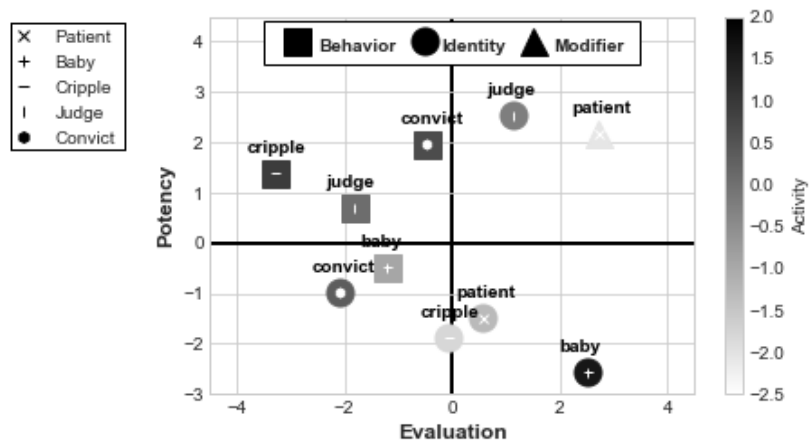

Figure 3: Visualization of words with different affective meaning in EPA space. Circles, squares, and triangles show identities, behaviors, and modifiers. The color represents the activity dimension. We can observe affective meaning of words such as "judge" is very different as an *identity* and a *behavior*.

Shallow word embeddings have only one representation for every word. On the other hand, the affective meaning of a word in *identity, modifier*, or *behavior* categories are different. For example, "mother", "coach", and "fool" have very different affective meanings when they are *behavior* or *identity*, but these words have only one representation in shallow embedding space. In fact, all the above approaches are limited by using one representation for different meanings of a word and not considering the context. Figure 3 shows the EPA values for some words that appear in different categories. We can observe how some words are mapped very differently based on their category. For example, "baby" as an *identity* is a pleasant and active character but is unpleasant and passive as a *behavior*.

To get a sense of how similar the affective meanings are between categories, we calculated the pairwise correlation of EPA values for words shared across various different categories (Table 1). Table 1a contains the correlations between the EPA values of words that appear as both an *identity* and a *modifier* in this dictionary. We see that for terms that can be either *identity* (a patient) or a *modifier* (a patient person), the correlation between their *evaluation* as a *modifier* and their *evaluation* as an *identity* is high (r=.93) but for *potency* it is less correlated (r=.62). Looking across Table 1, we see that affective meanings of the words in different appearing categories are not always highly correlated. For example, there is only a 0.4 correlation between *activity* dimensions of words that appear both in the *identity* and *behavior* categories. While some categories maintain high association across categories, other categories, like *identity* and *behavior* in the *activity* and *potency* dimensions have low association implying they words used in different categories represent different affective meanings, and consistent with the examples displayed in Figure 3. From Table 1, it is clear that the *activity* and *potency* sentiments associated with common words that can serve as either an *identity* or a *behavior* are too small to assume they represent the same affective meanings. This highlights the need for models that can represent contextual aspects and differentiate between different meanings of a word.

Table 1: Correlation among the affective meanings of identities, behaviors, and modifiers from a sentiment dictionary collected in 2014 (Smith-Lovin et al., 2016).

| (a) | **Identity-Modifier** | | | (b) | **Modifier-Behavior** | | | (c) | **Identity-Behavior** | | |
|---|---|---|---|---|---|---|---|---|---|---|---|
|   | **E** | **P** | **A** |   | **E** | **P** | **A** |   | **E** | **P** | **A** |
| **E** | 0.93*** | 0.49 | -0.58 | **E** | 0.98*** | 0.92*** | -0.32 | **E** | 0.73*** | 0.35 | 0.40 |
| **P** | 0.77* | 0.62 | -0.39 | **P** | 0.85** | 0.80* | -0.25 | **P** | 0.29 | 0.55* | 0.02 |
| **A** | -0.45 | 0.33 | 0.98*** | **A** | -0.27 | -0.30 | 0.67 | **A** | -0.11 | 0.30 | 0.40 |

Given the somewhat poorer performance of the shallow word-embedding approaches at predicting potency and activity and the hints in 1 about the lower correlations between sentiments across categories (e.g., identity and *behavior*; *modifier* and *behavior*, or *identity* and *modifier*) for the dimensions of *potency* and *activity*, we speculate that failure to consider context may be a contributing factor.

van Loon and Freese (2022) conducted four studies. In the first, they used "Word2Vec" in a word-embedding algorithm and a sentiment dictionary collected with an mTurk population (Smith-Lovin et al., 2019). In a second study, they used "GloVe" and "Word2Vec" word embeddings with a predictive modeling approach to estimate the same sentiment dictionary. In the second study, word embeddings play the role of features that are passed to a neural network to estimate sentiment dictionary. They saw a substantial improvement in sentiment predictions, particularly in the *activity* and *potency* dimensions. In a third study, they examined that same approach on newly collected data and found that it performed similarly. In a fourth study, however, they used a newer approach, capable of incorporating context more fully into the embeddings (BERT, described in more detail below). When they compared their models using BERT-embeddings on the same two datasets they received similar results to those in the second and third studies (combining Word2Vec and GloVe).

We think that pre-trained deep neural network has several advantages – including consideration of the contextual aspects of concepts within social events. We speculate that the failure to observe substantial improvements using BERT in van Loon and Freese (2022) may stem from the particular approach used in this study – which (a) excluded compound words or words such as "neighborly" that are not in BERT vocabulary. They extracted the meaning for word "roots" rather than concepts within events – aggregating across much of what makes BERT a context-embedding approach and (b) used one-hot encoding to define categories of concepts. While this approach still extracts the context-embedding of each root, it does not consider information about the event in a way that would be optimal from an affect control theory perspective.

## 1.4 Bidirectional Encoder Representations from Transformers

In 2018, Google open-sourced a language representation model named BERT as the state-of-the- art model for a wide range of Natural Language Processing (NLP) tasks. This deep neural network model pre-trained the bidirectional representation from a large set of unlabeled text. The model is pre-trained on from the BookCorpus and Wikipedia. BookCorpus includes 11,038 books with about 1000M words Zhu et al. (2015). English Wikipedia also has about 800M words.

Given a sentence, BERT first parses it into its parts. Then tokenizes the parsed sentence to transfer it to a sequence of tokens. Two special tokens are added at the beginning and end of the list. This sequence of tokens replaces the original sentence. This process is repeated for every sentence in the training set. The goal is to find high-dimensional vectors for every token in the training set such that the deep neural network can represent some language relationships. The loss function mathematically represents these relationships. The deep neural network maps these tokens to a high-dimensional numerical vector. Each of these high-dimensional vectors represents a token embedding, and the vector corresponding to one of the two additional tokens represents sentence embedding. We briefly review the process here to understand how BERT finds these vectors.

As a contextual representation, the BERT tokenizer, known as WordPiece, tokenizes the input sentences. If the sentence tokens are in the vocabulary of the pre-trained model, they appear in the tokenized list without any modification. However, WordPiece may assign multiple tokens to a word if the word is not in its vocabulary. In that case, tokens are root vocabulary and suffix of the original word. For example, if the words "affective" and "subtext" are not included in the vocabulary, WordPiece outputs [affect, ##ive] and [sub, ##text] tokens where ## shows the two tokens came from a compound word. Vocabulary of the pre-trained BERT model that we used includes 30,000 tokens and it can process many other compound words that are not in the vocabulary list. Indices of the tokens in the BERT tokenizer vocabulary are called token IDs.

After tokenization, the sentences are represented by a sequence of Token IDs. Since BERT is trained on "next sentence prediction" task, it assumes these sequences represent two sentences and two special tokens to indicate their relationship. The [CLS] token indicates the start of the first sentence, and the special token [SEP] comes at its end. We use BERT with only one sentence, but we have to use these special tokens. Figure 4 shows how a sentence representing a social interaction is tokenized and passed to the model (McCormick and Ryan, 2019). The output layer of the BERT model gives the embeddings for all the tokens shown as $C, T_1, ..., T_N, T_{[SEP]}$ where $T_k$ is the vector representation of $k^{th}$ token and $C$ corresponds to the $[CLS]$ token and can represent the sentence embedding.

The loss function in BERT model is defined for two objectives. (a) Masked language model and (b) next sentence prediction that we briefly review here. Masked language model objective is predicting some randomly masked tokens from a sentence. It uses context words on both sides of the target word in all layers to find the masked word. In other words, it randomly masks some words in a sentence and then uses the remaining contextual words to predict the masked words. We can describe this process by masking a word in the following sentence. *The student asked a question about the prerequisite courses.* Masking the word "question" we get,*The student asked a [MASK] about the prerequisite courses.* The model should use contextual words to predict this masked word. The ability to use full context in prediction differentiates BERT from other word embedding models like word2vec, which only uses the neighbor words for prediction. In addition to masked word prediction, BERT is also pre-trained on a "next sentence prediction" task. This task helps BERT better understand longer term relationship between sentences. As a result, pre-trained BERT can do great on tasks such as text summarization (Liu and Lapata, 2019). After fine-tuning for specific tasks, BERT gives state of the art in many challenging NLP tasks (McCormick and Ryan, 2019). We use the LARGE pre-trained version of BERT in this study and fine-tune it to find a numerical representation of a sentence that describes one social event.

In this project we used BERT large model (uncased) which is a 24-layer neural network, with 1024 hidden dimension and 16 attention heads. This model has 336M parameters (Devlin et al., 2018).

## 2   Methodology

The main advantage of word representation derived from BERT over shallow word-embeddings is that BERT can take into account the context of a word. This means that words can be given different representations when the words is used a as a *behavior* or an *identity* if we use them in a proper sentence. To take advantage of BERT embedding, we should train it on synthetic data that represents the concepts and their affective category simultaneously. In the analysis that follows, we show how to generate a contextual data-set describing social events to train a deep neural network and use this network to predict affective meanings of new concepts. We use BERT and fine tune it for finding affective lexicons. Our approach uses the vectorized representation of the whole sentence.

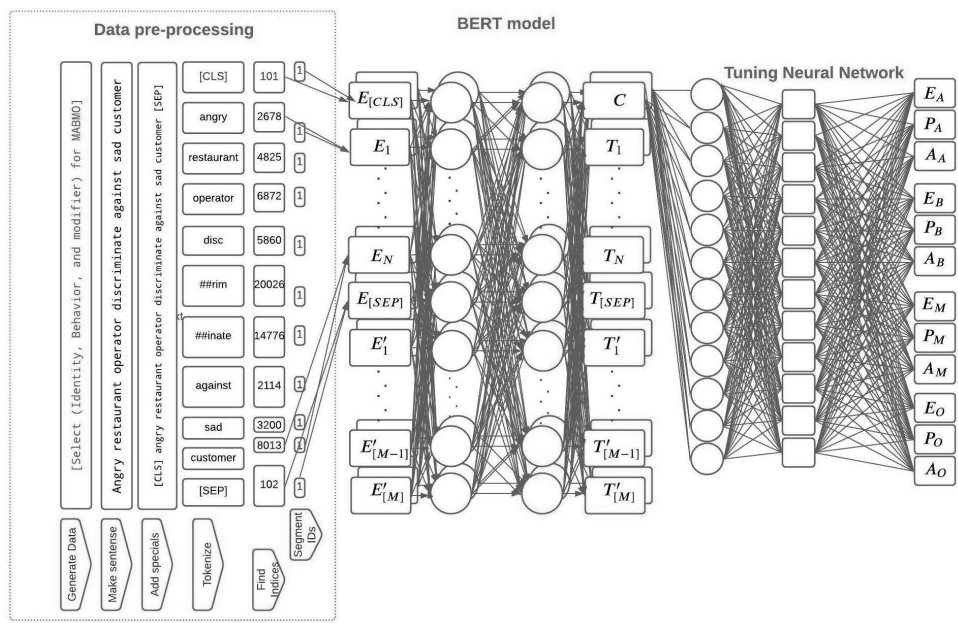

Figure 4: BERTNN pipeline. Sentences describing an event are generated in the first step and pre-processed to pass to a BERT model. Then outputs corresponding to the [CLS] token is passed to a three-layer neural network to find affective meaning.

We introduce a new framework, which processes synthetic data, passes it to a pre-trained BERT model, and fine-tunes the result by a three layer neural network to generate extended affective meaning dictionaries. The pipeline for our method (BERTNN) is shown in Figure 4 and we discuss the details in this section. The data used to train our model was generated from the affective dictionary described in Smith-Lovin et al. (2016). This dictionary was developed from surveys conducted between 2012-2014 and includes 929 *identities*, 814 *behaviors*, and 660 *modifiers*. The data used in this study is publicly available on the internet.

To take advantage of the BERT model, we need to make some synthetic data to fine-tune this model. This synthetic data is a set of sentences that use the concepts of interest. van Loon and Freese (2022) used token embedding of terms such as "assailant is" to find the embedding for "assailant". In this case, they get one estimation for each concept. In BERTNN, we use synthetic sentences that describe *Modified Actor Behaves Modified Actor* (MABMO) events and estimate the affective meaning of all concepts in the event. In this case, we use the vectorized representation of the whole sentence to estimate a 15-dimensional lexicon vector. This gives different estimations for a concept depending on other concepts used with the concept of interest in MABMO grammar. Then we aggregate multiple estimations to get a distribution for the concept of interest.

It takes only a few iterations to fine-tune the BERT model for affective meaning estimation in BERTNN. As a result, event selection is critical to get a fine-tuned model that is general enough to predict affective meaning for new concepts. We defined some pre-processing steps to select events that are general enough to span the diverse event space. The pre-processed data generate synthetic data that can represent the whole affective space of the dictionaries. Algorithm 1 describes the pre-processing algorithm to include contextual aspects of concepts before using BERT model.

Affective meaning of the concepts are distributed across the EPA dimensions. However, they are not uniformly distributed across all three dimensions. Since we use a few iterations to fine-tune BERT model, random sampling from these values may use a train set from a specific region that is not necessarily the whole space. For example, all the training samples may have positive activity for the behavior and as a result, the model could be more accurate if we have samples with negative behavior. Every concept in *identity*, *behavior*, and *modifier* dictionaries is represented by a three dimensional affective vector. We are looking for training samples that represents different regions/clusters within each dictionary. In the first step of Algorithm 1, we needed to know how many different regions/clusters we have in the dictionaries and uniformly sample from

**Data:** Affective dictionaries for Identities, Modifiers, and Behaviors
1. Find clusters/regions for the concepts in each of the Identity, Modifier, and Behavior dictionaries
2. Split the data into train, test, and validation sets using stratified sampling on the clusters
3. Make all ABO events using the train set
4. Use impression change equations to find the impression of concepts in all events.
5. Find the sign of the difference between baseline sentiments and impressions of the events across all 9 dimensions ( 3 concepts in ABO grammer $\times$ 3 EPA dimensions). In other words, for the actor, the behavior, and the object, we find whether the impression of the event increases or decreases each of their EPA values.
6. Use binary encoding to label variations of the events. For example, 111000111 means that the impression of the event for the actor and the object are increased across all EPA dimensions and decreases for the behavior across all EPA dimensions.
7. Add two random modifiers to ABO events created in the previous step to make synthetic MABMO events.
8. Define a training data generator that uses defined binary encodings to make synthetic data that spans all possible different interactions.

**Algorithm 1:** Pre-processing algorithm

each of them. For this purpose, we clustered each *identity*, *behavior*, and *modifier* dictionaries using K-Means clustering. We used the elbow method and determined the number of clusters for each of these three dictionaries. We concluded that 5 clusters/regions are enough for all three categories.

As a common practice in data-mining methods, the second step in Algorithm 1 is to partition the data into training, test, and validation set. We used the train set to fine-tune the BERT model and the test set to choose hyper-parameters such as the number of epochs. We selected hyper-parameters that resulted in an acceptable loss value for the test set. The validation set was the hold-out data. The results reported in this paper come from the validation set. We used stratified sampling on regions/clusters in affective dictionaries to get similar distributions. Each cluster of *identity*, *behavior*, and *modifier* categories are one "strata". We randomly sample 85% of the words in each strata for the training, 8% are selected as the test set, and the remaining 12% are used as a validation set.

As we discussed the importance of *deflection* earlier, in the third step, we are looking to capture variation in factors that change deflection. When we select some events to train the model, they should represent the variation of concept impression. The training data should include events that increase or decrease the baseline sentiments in each dimension. For the train set, we found the impression of all possible events in ABO grammar using impression change equations. Then the difference between the impressions and baseline concepts are calculated across all 9 dimensions. Among all $2^9 = 512$ possible variations, 508 type of events were present based on the training set. We used binary encoding to label these 508 types of events.

Affective dictionaries are collected from surveys that participants rate a concept in one of the *identity, behavior,* or *modofier* categories. On the other hand, to utilize a BERT model we should make synthetic data that is a list of sentences that represents words of interests and their corresponding category. To generate each synthetic sentence, we used two *identities*, two *modifiers*, and one *behavior* from the training set. Synthetic Sentences with an MABMO grammar (e.g., Happy employee greets bossy employer) represent a social event. In other words, each sample describes an event in MABMO grammar.

Algorithm 1 labels ABO events. For MABMO grammar, we need to sample two modifiers and one ABO event. We used stratified sampling based on modifier clusters and ABO labels developed in Algorithm 1. After pre-processing and tokenizing MABMO sentences similar to "data pre-processing" part of the pipeline, they are sent to the pre-trained BERT model.

There are two main approaches to getting vector representations from BERT. One is using the token embedding for the tokens of a word. This method is not efficient when we have compound words or we are dealing with a word outside the BERT dictionary. Alternatively, the embedding for the [CLS] token can represent an embedding for the whole sentence. Using fine-tuning layers, it is possible to get representations for the words from this sentence embedding. We used BERT outputs that correspond to [CLS] token as a vectorized sentence representation. The next step is finding a mapping from the [CLS] outputs to the affective space. We passed BERT output to a fine-tuning neural network with a dense input layer followed by a "Relu" hidden layer. Relu function adds required non-linearity in the network. The last layer is a dense layer that produces the 15-dimensional vector representing EPA values for all MABMO characters. The $L_2$ loss used in the neural networks minimized the squared error between estimated affective meaning and the target values across all 15 dimensions.

We implemented the neural network in Python using Torch package. We used "AdamW" with a learning rate of 2e-5 and batch size of 64. The BERT model and neural network are tuned for a few epoch. The data in the testing set is used to decide on how many iteration results in a good model. The neural network output is a 15-dimension vector that is highly correlated with the affective meaning variables. The code is available upon request to the first author.

Table 2: Comparing four different affective dictionaries (Indiana (Francis and Heise, 2006), Texas (Schneider, 2006), and North Carolina (Smith-Lovin and Heise, 1978)) with the affective dictionary used in this study (Smith-Lovin et al., 2016).

|  |  | MAD | | | RMSD | | | Correlation | | | Count |
|---|---|---|---|---|---|---|---|---|---|---|---|
|  |  | E | P | A | E | P | A | E | P | A |  |
| Identity | Indiana | 0.41 | 0.48 | 0.44 | 0.53 | 0.63 | 0.57 | 0.96*** | 0.9*** | 0.81*** | 317 |
|  | NorthCarolina | 0.57 | 0.58 | 0.69 | 0.71 | 0.71 | 0.85 | 0.93*** | 0.91*** | 0.65*** | 453 |
|  | Texas | 0.45 | 0.52 | 0.77 | 0.59 | 0.64 | 0.95 | 0.94*** | 0.91*** | 0.54*** | 319 |
| Behavior | Indiana | 0.54 | 0.61 | 0.59 | 0.67 | 0.74 | 0.73 | 0.97*** | 0.8*** | 0.67*** | 288 |
|  | NorthCarolina | 0.68 | 0.5 | 0.52 | 0.82 | 0.63 | 0.65 | 0.95*** | 0.8*** | 0.7*** | 500 |
|  | Texas | 0.53 | 0.56 | 0.76 | 0.65 | 0.69 | 0.94 | 0.96*** | 0.81*** | 0.32*** | 225 |
| Modifier | Indiana | 0.61 | 0.52 | 0.58 | 0.71 | 0.65 | 0.7 | 0.97*** | 0.92*** | 0.86*** | 276 |
|  | NorthCarolina | 0.78 | 0.72 | 0.63 | 0.92 | 0.89 | 0.81 | 0.96*** | 0.91*** | 0.79*** | 407 |
|  | Texas | 0.78 | 0.6 | 0.78 | 0.88 | 0.73 | 0.92 | 0.97*** | 0.91*** | 0.79*** | 58 |

Table 3: Summary statistics of estimated EPA values for "jeweler" as an actor.

|  | EEA | EPA | EAA | E | P | A |
|---|---|---|---|---|---|---|
| Mean | 1.33 | 1 | 0.04 | 0.79 | 0.49 | -0.63 |
| S.D. | 0.24 | 0.2 | 0.19 | 0.94 | 1.32 | 1.34 |
| Min | 0.61 | 0.4 | -0.59 | -0.7 | -4.3 | -3.6 |
| Max | 2.02 | 1.62 | 0.85 | 3.6 | 2.5 | 2.1 |

The tuning layers in our framework are trained based on the target affective values. In training the neural network, we used early stopping to get reasonably low errors in the training and test set. To decide on reasonably low thresholds, we compared Mean Absolute Distance (MAD), Root mean Square Distance (RMSD), and the correlation between common items in different affective dictionaries. Table 2 compares the affective dictionary used in this study (Smith-Lovin et al., 2016) with other affective dictionaries.

We actively checked L2loss for the training and test set and had early stopping considering both values. We stopped the process after we used 247 batches of size 64 to train the model. After convergence, the trained model can predict the affective meaning for a large number of concepts. Since we have a contextual representation of words, we get multiple representations for a word from BERTNN depending on other terms in MABMO event. In other word, for affective meaning of a given *actor* in MABMO grammar, we get different estimates depending other *modifiers, behavior*, and *object* used in the event. For an example of this, Table 3 shows variation of values we get for "jeweler" in the test set,

From our model, we sampled 30K events and found 385 events that had "jeweler" as the actor. In Table 3, estimated values generated from these 385 samples in test set is shown on the left side. The left columns in the table show the estimation from BERTNN in the test set. The right three columns show the EPA sentiments from 61 samples containing this word in the original survey ratings. We can observe that estimated evaluation dimensions varied from 0.61 to 2.02. On the other hand, it varied even more widely in the survey data – from -.7 to 3.6. So the variation of estimated values from our model is smaller than variations among different participants in the original study. We observe similar trends for most of the words in the dictionary.

Since the BERT model returns multiple representations for a concept, we have to adopt an aggregation approach. One approach is generating multiple MABMO events including the concept of interest. Then the estimated affective meaning values in these events represent a distribution for its value. We can use the mean value as an estimation. For example, if "moderator" is a new *identity* outside the affective dictionary, we make multiple events that includes this identity as the *actor* such as, "moderator help angry client". All events are passed to the model which estimates the affective meaning of the target word in each event. The average affective meaning is returned as the final estimated meaning.

Comparing BERTNN with (van Loon and Freese, 2022) work, the main differences are,

- BERTNN makes multiple synthetic sentences to take advantage of contextual embedding of the BERT model. These sentences are made based on MABMO grammar in ACT literature. It helps BERTNN to consider deflection and regions of concepts in training.

- Instead of optimizing only the fine-tuning layer, BERTNN trains the BERT+fine-tuning layer for a few epochs.

- BERTNN uses sentence embedding instead of the token embedding of the core concepts. As a result, it can estimate affective meaning for compound words and words with more than one token.

- BERTNN gives a distribution for affective meaning based on the synthetic data used.

We make BERTNN a publicly available tool that researchers with minimal coding experience can use to estimate the affective meaning of their words of interest. This tool is available in the following Google Collab link,

https://colab.research.google.com/drive/1ej1wldgDgjOOu2OBf3xXasq51L6V-gft?usp=sharing.

## 3  Results

Variations in estimated affective meanings from the model can arise from a variety of sources, including,

- Splits of the original data to make train, test, and validation changes the final result. Consequently, we conducted a series of robustness checks. We get similar results using the test set for model selection and increasing or decreasing the number of iterations. We also simulated this process with various seeds, and the final results are similar.

Table 4: Performance of several models on (Smith-Lovin et al., 2016) data. Bold indicates the best model. Our model, BERTNN, performed best in most categories.

|  |  | MAE | | | RMSE | | | Correlation | | |
|---|---|---|---|---|---|---|---|---|---|---|
|  |  | E | P | A | E | P | A | E | P | A |
| **Identity** | Analogy stepW. | 0.93 | 0.92 | 0.81 | 1.2 | 1.13 | 0.99 | 0.65 | 0.53 | 0.18 |
|  | Analogy_regression | 0.95 | 0.95 | 0.81 | 1.22 | 1.16 | 1.00 | 0.64 | 0.48 | 0.15 |
|  | StepW Translation (Mostafavi and Porter, 2021) | 0.74 | 0.75 | 0.67 | 0.97 | 1.04 | 0.87 | 0.82 | 0.64 | 0.50 |
|  | CoreBERT (van Loon and Freese, 2022) | 0.70 | 0.65 | 0.66 | 0.93 | 0.87 | 0.82 | 0.86 | 0.71 | 0.53 |
|  | BERTNN | **0.55** | **0.57** | **0.51** | **0.77** | **0.76** | **0.68** | **0.89** | **0.81** | **0.68** |
| **Behavior** | Analogy stepW. | 1.17 | 0.71 | 0.68 | 1.44 | 0.90 | 0.83 | 0.73 | 0.45 | 0.51 |
|  | Analogy_regression | 1.20 | 0.75 | 0.73 | 1.48 | 0.93 | 0.90 | 0.71 | 0.44 | 0.38 |
|  | StepW Translation (Mostafavi and Porter, 2021) | 0.95 | 0.67 | 0.64 | 1.21 | 0.84 | 0.82 | 0.80 | 0.52 | 0.52 |
|  | CoreBERT (van Loon and Freese, 2022) | **0.69** | 0.48 | **0.46** | **0.90** | 0.61 | **0.60** | **0.90** | 0.75 | 0.77 |
|  | BERTNN | 0.73 | **0.43** | 0.49 | 0.96 | **0.57** | 0.61 | 0.87 | **0.80** | **0.79** |
| **Modifier** | Analogy stepW. | 0.33 | 0.73 | 0.88 | 1.08 | 0.9 | 1.09 | 0.87 | 0.86 | 0.57 |
|  | Analogy_regression | 0.56 | 0.76 | 0.92 | 1.16 | 0.93 | 1.10 | 0.86 | 0.86 | 0.57 |
|  | StepW Translation (Mostafavi and Porter, 2021) | 0.79 | 0.48 | 0.66 | 0.98 | 0.67 | 0.84 | 0.90 | 0.89 | 0.74 |
|  | CoreBERT (van Loon and Freese, 2022) | 0.65 | 0.53 | 0.66 | 0.82 | 0.66 | 0.83 | 0.92 | 0.89 | 0.77 |
|  | BERTNN | **0.57** | **0.48** | **0.50** | **0.74.** | **0.59** | **0.62** | **0.94** | **0.91** | **0.87** |

- Mean, median, or similar statistics such as trim-mean of multiple sample estimations for a concept, result in different final results. In this study, we used the mean values across all dimensions.

- An *identity* is used as either actor or *object* in the MABMO grammar. Similarly, *modifier* can modify actor or *object*. The values that we get for a concept in these cases varies but they are close. We considered concatenating the results and aggregating for the concept in general. But we consider one of the strengths of our chosen approach as the ability to distinguish different affective meanings for concepts operating in the role of an *actor* or *object*. For simplicity of the model, we used *actor* to estimate *identity* and its *modifier* to estimate *modifiers*. It is possible to run studies to differentiate the values one can get for *identity* as the *actor* or the *object*.

We compared the performance of our model with approaches tried in previous work such as different word analogy (Kozlowski et al., 2019), regressions (Li et al., 2017), translation matrix methods (Mostafavi and Porter, 2021), and BERT model on the core concepts (van Loon and Freese, 2022) to find affective meanings in Table 4. We used RMSE, MAE, and correlation analysis to compare the result. The validation data used to compare these methods included 139 *identities*, 122 *behaviors*, and 99 *modifiers* came from stratified sampling discussed earlier.

Table 5: Comparing estimated affective meanings for "judge" as an *identity* and *behavior*.

| *Judge* | Evaluation | Potency | Activity | Est. Evaluation | Est. Potency | Est. Activity |
|---|---|---|---|---|---|---|
| **Behavior** | -1.83 | 0.71 | 0.07 | -1.91 | 0.73 | 0.08 |
| **Identity** | 1.15 | 2.53 | -0.22 | 1.18 | 2.54 | -0.17 |

All the similar works except (van Loon and Freese, 2022) used shallow embedding (Kozlowski et al., 2019; Mostafavi and Porter, 2021; Li et al., 2017). We used the publicly available code of Mostafavi and Porter (2021) and Kozlowski et al. (2019) to replicate their work and compare the result. For Li et al. (2017) we had to implement their method in Python. To further improve their methods, we added tuning layers such as adding step-wise regression to the analogy method. We used pre-trained model in van Loon and Freese (2022) to compare with their work. Table 4, shows the best result we could get from other methods. We can observe from this table that our approach reached the best result across most of the metrics. We can observe from Table 4 that in some metrics such as the correlation metric for Activity, the improvement over shallow-embedding methods is about 20% for identities and behaviors. In terms of error rates, we can find BERTNN achieves values that are comparable with differences between different affective dictionaries. For example, MAE value that BERTNN achieves for modifiers in Table 4 is smaller than three of dictionaries shown in Table 2. In other word, BERTNN estimation error on the validation set is small enough to say this model estimates EPA values similar to running a new survey.

One problem with estimation from shallow-embeddings was having the same embedding for words as *identity* or *behaviors*. Using the BERTNN, we can differentiate between these two cases. In In Table 5 we can observe that estimated values for "judge" are different when it is considered as *identity* or *behavior*.

To evaluate how close our expanded dictionaries are to the baseline affective meanings, we calculated the correlation for *identity* and *behaviors* in Tables 6 and 7. The correlation (a) between the result of the validation set in our method, (EE, EP, and EA) and the EPA values from the surveys (E, P, and A) are shown. The diagonal terms are the correlation values we have seen earlier in Table 4. Also, you can find the correlation between the three dimensions from the survey data are shown in (b), and the correlation between estimated three dimensions are shown in (c). We can observe in both tables the values in tables (b) and (c) are close. It reveals that BERTNN estimation is highly correlated with survey data and the cross-term dynamics are well estimated.

Table 6: Correlation analysis for identities. (a) Correlation between estimated values (EE, EP, and EA) and the affective dictionary value (E, P, and A). We can also compare the correlation of the words in the three dimensions shown in (b) and correlation of the estimated values of the three dimension shown in (c) to find how close the off-diagonal entries are in the estimation comparing to dictionary values.

| (a) | E | P | A | | (b) | E | P | A | | (c) | EE | EP | EA |
|---|---|---|---|---|---|---|---|---|---|---|---|---|---|
| EE | 0.89 | 0.56 | -0.08 | | E | 1.00 | 0.55 | 0.05 | | EE | 1.00 | 0.62 | 0.04 |
| EP | 0.59 | 0.81 | 0.29 | | P | 0.55 | 1.00 | 0.25 | | EP | 0.62 | 1.00 | 0.29 |
| EA | 0.02 | 0.15 | 0.68 | | A | 0.05 | 0.25 | 1.00 | | EA | 0.04 | 0.29 | 1.00 |

Table 7: Correlation analysis for behaviors. (a) Correlation between estimated values (EA, EP, and EA) and the affective dictionary value (E, P, and A). We can also compare the correlation of the words in the three dimensions shown in (b) and correlation of the estimated values of the three dimension shown in (c) to find how close the off-diagonal entries are in the estimation comparing to dictionary values.

| (a) | E | P | A | | (b) | E | P | A | | (c) | EE | EP | EA |
|---|---|---|---|---|---|---|---|---|---|---|---|---|---|
| EE | 0.87 | 0.38 | -0.37 | | E | 1.00 | 0.59 | -0.21 | | EE | 1.00 | 0.47 | -0.42 |
| EP | 0.36 | 0.80 | 0.15 | | P | 0.59 | 1.00 | 0.12 | | EP | 0.47 | 1.00 | 0.12 |
| EA | -0.33 | 0.09 | 0.79 | | A | -0.21 | 0.12 | 1.00 | | EA | -0.42 | 0.12 | 1.00 |

Tables 6 and 7 show that diagonal terms in the correlation of estimated values and values from the survey dictionary are reasonably large. On the other hand, the off-diagonal entries from the estimation are close to the ones from the survey.

**Model outputs.** BERTNN can represent one participant evaluating affective meaning of concepts from an MABMO grammar. Table 8 shows estimation of concepts used in *Happy doctor help wonderful mother*.

Since BERTNN gives us contexal representation of words, we can create multiple MABMO events with a common word to estimate its affective meaning in general. For example, *fight* as a behavior is a concept that is not evaluated in affective dictionaries. We can create multiple events similar to estimate its affective meaning in different contexts. Table 9 includes estimated affective meaning from multiple MABMO events. Agrregating 300 events similar to Table 9, we can get a reliable estimate for the concept of interest. Table 10 has a summary statistics of events used to estimate affective meaning of *fight* as a behavior.

Table 8: Using BERTNN to represent one participant evaluating concepts in MABMO grammar.

| Modifier | | | Actor | | | Behavior | | | Modifier | | | Object | | |
|---|---|---|---|---|---|---|---|---|---|---|---|---|---|---|
| Happy | | | doctor | | | help | | | wonderful | | | mother | | |
| E | P | A | E | P | A | E | P | A | E | P | A | E | P | A |
| 2.09 | 2.12 | 0.54 | 1.81 | 2.03 | 0.34 | 2.3 | 2.65 | 0.7 | 2.23 | 2.16 | 0.39 | 2.23 | 2.11 | 0.41 |

Table 9: Estimated affective meaning of *fight* from MABMO different events.

| MABMO event | E | P | A |
|---|---|---|---|
| Decisive outsider fight unmotivated cutthroat | -1.98 | 1.02 | 2.49 |
| Peaceful decorator fight amused cement worker | -1.94 | 0.99 | 2.59 |
| Deprave tease fight panicked prostitute | -1.98 | 1.02 | 2.65 |
| Artistic pal fight disheartened gangster | -2.14 | 1.03 | 2.60 |
| Old fashioned casual laborer fight petty undergraduate | -2.11 | 0.95 | 2.63 |
| Sloppy astrologer fight indifferent celebrity | -1.98 | 0.96 | 2.55 |

Table 10: Estimating affective meaning of *fight* from 300 MABMO events.

|  | E | P | A |
|---|---|---|---|
| Mean | -2.07 | 1.02 | 2.60 |
| Standard deviation | 0.12 | 0.04 | 0.07 |
| Minimum | -2.36 | 0.92 | 2.20 |
| Maximum | -1.29 | 1.21 | 2.76 |

## 4 Discussion

Creating a synthetic data set of events corresponding to the core grammar of ACT, pre-processing these data and then sending them through the pre-trained BERT word embedding model yielded predictions that seem to have great promise for generating the sort of cultural sentiments required by ACT. We argue that this approach produces sentiments that are closer to the concept sentiments required for ACT, rather than sentiments for words or root words. We also note that the learning approach used by BERTNN is not unlike the means by which humans acquire social meaning through the observation of participation in social interaction. When we encounter a new culture or subculture with an identity we do not recognize, we observe interactions in order to gather information. If we observe occupants of this identity in many interactions (some of the form MABMO), each interaction leaves an impression about the likely meaning of that identity. If someone always does good things, we will come to infer a positive evaluation for his/her identity. So if we don't know the identity of an actor (in this context, the identity of a person is masked in MABMO interactions), we would have something similar to M*BMO. Based on 500 interactions, an affective meaning for the unknown identity is formed in our minds. Like the impression change equations we described previously, we expect the person to be nice if we observe lots of nice behaviors in observed interactions, or routinely interacts with other positively identified actors. In this way, our method of learning the affective meanings associated with known/unknown identities is very similar to the mask language model.

ACT is one of sociology's most enduring and developed mathematical theories. Nonetheless, application and extensions of the theory are seriously limited by the need for resource-intensive data collection order to incorporate concepts from new substantive domains. Traditional methods require time, expense, and significant burden on human research participants and researchers. Machine learning approaches of the sort described in this paper promise to minimize this burden by sidestepping the need for human raters. This approach could greatly expand the ability of ACT to be applied to new substantive domains at greater speed and at a lower burden for research respondents, researchers, and funders. Using machine learning approaches pre-trained on very large data sets allow the possibility of rapidly expanding the existing sentiment dictionaries "on the fly" rather than needing to design, implement, analyze, and interpret new cultural data. Moreover, the ability to train these context-embedding models on new data sets opens up the possibility of moving into novel cultural and subcultural domains more quickly (skipping years of fieldwork or thousands of surveys).

# References

Alhothali, Areej and Jesse Hoey. 2017. "Semi-supervised affective meaning lexicon expansion using semantic and distributed word representations." *arXiv preprint arXiv:1703.09825* .

Averett, Christine and David R Heise. 1987. "Modified social identities: Amalgamations, attributions, and emotions." *Journal of Mathematical Sociology* 13:103–132.

Britt, Lory and David R Heise. 1992. "Impressions of self-directed action." *Social Psychology Quarterly* pp. 335–350.

Devlin, Jacob, Ming-Wei Chang, Kenton Lee, and Kristina Toutanova. 2018. "Bert: Pre-training of deep bidirectional transformers for language understanding." *arXiv preprint arXiv:1810.04805* .

Fontaine, Johnny RJ, Klaus R Scherer, Etienne B Roesch, and Phoebe C Ellsworth. 2007. "The world of emotions is not two-dimensional." *Psychological science* 18:1050–1057.

Francis, Clare and David R Heise. 2006. "Mean affective ratings of 1,500 concepts by Indiana University undergraduates in 2002–3, 2006." *Computer file]. Distributed at Affect Control Theory Website, Program Interact (http://www. indiana. edu/~ socpsy/ACT/interact/JavaInteract. html)* .

Francis, Linda E. 1997a. "Emotion, coping, and therapeutic ideologies." *Social perspectives on emotion* 4:71–102.

Francis, Linda E. 1997b. "Ideology and interpersonal emotion management: Redefining identity in two support groups." *Social Psychology Quarterly* pp. 153–171.

Working paper - 2022                                                                                         25
Goldstein, David M. 1989. "Control theory applied to stress management." In *Advances in Psychology*, volume 62, pp. 481–491. Elsevier.

Heise, David R. 1977. "Social action as the control of affect." *Behavioral Science* 22:163–177.

Heise, David R. 2010. *Surveying cultures: Discovering shared conceptions and sentiments*. John Wiley & Sons.

Heise, David R. 2013. "Interact guide." *Department of Sociology, Indiana University* .

Heise, David R and Cassandra Calhan. 1995. "Emotion norms in interpersonal events." *Social Psychology Quarterly* pp. 223–240.

Heise, David R and Lisa Thomas. 1989. "Predicting impressions created by combinations of emotion and social identity." *Social Psychology Quarterly* pp. 141–148.

Hunt, Pamela M. 2008. "From festies to tourrats: Examining the relationship between jamband subculture involvement and role meanings." *Social Psychology Quarterly* 71:356–378.

Kozlowski, Austin C, Matt Taddy, and James A Evans. 2019. "The geometry of culture: Analyzing the meanings of class through word embeddings." *American Sociological Review* 84:905–949.

Kriegel, Darys J, Muhammad Abdul-Mageed, Jesse K Clark, Robert E Freeland, David R Heise, Dawn T Robinson, Kimberly B Rogers, and Lynn Smith-Lovin. 2017. "A multilevel investigation of Arabic-language impression change." *International Journal of Sociology* 47:278–295.

Li, Minglei, Qin Lu, Yunfei Long, and Lin Gui. 2017. "Inferring affective meanings of words from word embedding." *IEEE Transactions on Affective Computing* 8:443–456.

Liu, Yang and Mirella Lapata. 2019. "Text summarization with pretrained encoders." *arXiv preprint arXiv:1908.08345* .

MacKinnon, Neil J and Dawn T Robinson. 2014. "Back to the future: 25 years of research in affect control theory." *Advances in group processes* .

McCormick, Chris and Nick Ryan. 2019. "BERT Word Embeddings Tutorial."

Mikolov, Tomas, Kai Chen, Greg Corrado, and Jeffrey Dean. 2013. "Efficient estimation of word representations in vector space." *arXiv preprint arXiv:1301.3781* .

Mostafavi, Moeen. 2021. "Adapting Online Messaging Based on Emotional State." In *Proceedings of the 29th Conference on User Modeling, Adaptation and Personalization*.

Mostafavi, Moeen and Michael Porter. 2021. "How emoji and word embedding helps to unveil emotional transitions during online messaging." In *2021 IEEE International Systems Conference (SysCon)*. IEEE.

Osgood, Charles Egerton, William H May, Murray Samuel Miron, and Murray S Miron. 1975. *Cross-cultural universals of affective meaning*, volume 1. University of Illinois Press.

Osgood, Charles Egerton, George J Suci, and Percy H Tannenbaum. 1957. *The measurement of meaning*. University of Illinois press.

Rashotte, Lisa Slattery. 2002. "Incorporating nonverbal behaviors into affect control theory." *Electronic Journal of Sociology* 6.

Robillard, Julie M and Jesse Hoey. 2018. "Emotion and motivation in cognitive assistive technologies for dementia." *Computer* 51:24–34.

Robinson, Dawn T, Jody Clay-Warner, Christopher D Moore, Tiffani Everett, Alexander Watts, Traci N Tucker, and Chi Thai. 2012. "Toward an unobtrusive measure of emotion during interaction: Thermal imaging techniques." In *Biosociology and neurosociology*. Emerald Group Publishing Limited.

Robinson, Dawn T and Lynn Smith-Lovin. 1992. "Selective interaction as a strategy for identity maintenance: An affect control model." *Social Psychology Quarterly* pp. 12–28.

Robinson, Dawn T and Lynn Smith-Lovin. 1999. "Emotion display as a strategy for identity negotiation." *Motivation and Emotion* 23:73–104.

Robinson, Dawn T and Lynn Smith-Lovin. 2018. "Affect control theories of social interaction and self." *Contemporary social psychological theories* .

Robinson, Dawn T, Lynn Smith-Lovin, and Olga Tsoudis. 1994. "Heinous crime or unfortunate accident? The effects of remorse on responses to mock criminal confessions." *Social Forces* 73:175–190.

Rogers, Kimberly B. 2018. "Do you see what I see? Testing for individual differences in impressions of events." *Social Psychology Quarterly* 81:149–172.

Rogers, Kimberly B and Lynn Smith-Lovin. 2019. "Action, interaction, and groups." *The Wiley Blackwell Companion to Sociology* pp. 67–86.

Schneider, Andreas. 2006. "Mean Affective Ratings of 787 Concepts by Texas Tech University Undergraduates in 1998." *Distributed at UGA Affect Control Theory Website: http://research. franklin. uga. edu/act* .

Schröder, Tobias and Wolfgang Scholl. 2009. "Affective dynamics of leadership: An experimental test of affect control theory." *Social Psychology Quarterly* 72:180–197.

Smith, Herman W. 2002. "The dynamics of Japanese and American interpersonal events: Behavioral settings versus personality traits." *Journal of Mathematical Sociology* 26:71–92.

Smith, Herman W and Linda E Francis. 2005. "Social vs. self-directed events among Japanese and Americans." *Social Forces* 84:821–830.

Smith-Lovin, Lynn. 1979. "Behavior settings and impressions formed from social scenarios." *Social Psychology Quarterly* pp. 31–43.

Smith-Lovin, Lynn and William Douglass. 1992. "An affect control analysis of two religious subcultures." *Social perspectives on emotion* 1:217–47.

Smith-Lovin, Lynn and David R Heise. 1978. "Mean affective ratings of 2,106 concepts by University of North Carolina undergraduates in 1978 [computer file]."

Smith-Lovin, Lynn, Dawn T Robinson, Bryan C Cannon, Jesse K Clark, Robert Freeland, Jonathan H Morgan, and Kimberly B Rogers. 2016. "Mean affective ratings of 929 identities, 814 behaviors, and 660 modifiers by university of georgia and duke university undergraduates and by community members in durham, nc, in 2012-2014." *University of Georgia: Distributed at UGA Affect Control Theory Website: http://research. franklin. uga. edu/act* .

Smith-Lovin, Lynn, Dawn T Robinson, Bryan C Cannon, Brent H Curdy, and Jonathan H Morgan. 2019. "Mean affective ratings of 968 identities, 853 behaviors, and 660 modifiers by amazon mechanical turk workers in 2015." *University of Georgia: Distributed at UGA A ect Control eory Website* .

Tsoudis, Olga and Lynn Smith-Lovin. 1998. "How bad was it? The effects of victim and perpetrator emotion on responses to criminal court vignettes." *Social forces* 77:695–722.

van Loon, Austin and Jeremy Freese. 2022. "Word Embeddings Reveal How Fundamental Sentiments Structure Natural Language." *American Behavioral Scientist* .

Youngreen, Reef, Bridget Conlon, Dawn T Robinson, and Michael J Lovaglia. 2009. "Identity maintenance and cognitive test performance." *Social Science Research* 38:438–446.

Zhu, Yukun, Ryan Kiros, Rich Zemel, Ruslan Salakhutdinov, Raquel Urtasun, Antonio Torralba, and Sanja Fidler. 2015. "Aligning books and movies: Towards story-like visual explanations by watching movies and reading books." In *Proceedings of the IEEE international conference on computer vision*, pp. 19–27.